\documentclass{article}

\PassOptionsToPackage{numbers,sort&compress}{natbib}
\usepackage[preprint]{neurips_2025}
\usepackage[utf8]{inputenc} 
\usepackage[T1]{fontenc}    
\usepackage{hyperref}       
\usepackage{url}            
\usepackage{booktabs}       
\usepackage{amsfonts}       
\usepackage{nicefrac}       
\usepackage{microtype}      
\usepackage{xcolor}         
\usepackage{graphicx}
\usepackage{amsmath}
\usepackage{enumitem}
\usepackage{multirow}

\usepackage{listings}
\usepackage{tcolorbox}
\usepackage{amsmath}

\tcbuselibrary{listings,skins}
\newtcblisting{promptbox}[3][]{
    listing only,
    colback=gray!5,
    colframe=gray!30,
    coltitle=black,
    listing options={
        basicstyle=\footnotesize\ttfamily,
        breaklines=true,
        breakatwhitespace=true,
        columns=flexible,
        keepspaces=true,
        showstringspaces=false,
    },
    title=#2,
    label=#3,
    #1
}

\title{Symphony: A Decentralized Multi-Agent Framework for Scalable Collective Intelligence}

\author{
  Ji Wang\textsuperscript{1, 3},
  Kashing Chen\textsuperscript{1, 4},
  Xinyuan Song\textsuperscript{2},
  Ke Zhang\textsuperscript{1, 5},
  Lynn Ai\textsuperscript{1},
  Eric Yang\textsuperscript{1},
  Bill Shi\textsuperscript{1}\thanks{Corresponding author: tianyu@gradient.network} \\
  \textsuperscript{1} Gradient \\
  \textsuperscript{2} Dpartment of Computer Science, Emory University \\
  \textsuperscript{3} Department of Industrial Engineering and Oprations Research, Columbia University \\
  \textsuperscript{4} The Chinese University of Hong Kong \\
  \textsuperscript{5} Department of Computer Science and Computer Engineering, Waseda University \\
}

\begin{document}

\maketitle

\begin{abstract}

Most existing Large Language Model (LLM)-based agent frameworks rely on centralized orchestration, incurring high deployment costs, rigid communication topologies, and limited adaptability. To address these challenges, we introduce \textbf{Symphony}, a decentralized multi-agent system which enables lightweight LLMs on consumer-grade GPUs to coordinate. Symphony introduces three key mechanisms: (1) a decentralized ledger that records capabilities, (2) a Beacon-selection protocol for dynamic task allocation, and (3) weighted result voting based on CoTs. This design forms a privacy-saving, scalable, and fault-tolerant orchestration with low overhead. Empirically, Symphony outperforms existing baselines on reasoning benchmarks, achieving substantial accuracy gains and demonstrating robustness across models of varying capacities. Our code is available at \url{https://github.com/GradientHQ/Symphony.git}.
\end{abstract}

\section{Introduction}

Large Language Models (LLMs) have demonstrated strong performance in natural language understanding, reasoning~\cite{zhu20254th}, planning~\cite{zhang2025ccma}, and tool use~\cite{toubal2024modeling}, powering diverse domains, such as education~\cite{zhang2024simulating}, healthcare~\cite{chu2023age}, autonomous systems~\cite{yang2024wcdt}, software engineering~\cite{hou2023large}, and scientific discovery~\cite{wang2023sample}. Driven by LLMs, machine learning systems are increasingly dependent on agent-based orchestration to address complicated tasks. Resent frameworks such as AutoGen~\cite{wu2023autogen}, MetaGPT~\cite{metagpt2023}, CAMEL~\cite{camel2023}, Voyager~\cite{wang2023voyager}, and CrewAI~\cite{crewai} have demonstrated that agent cooperation enhances task solving. However, most LLM-based systems adopt a centralized architecture where a single agent manages task allocation, orchestrates message routing, and monitors whole workflow, resulting in scalability bottlenecks~\cite{aratchige2025llms}, rigid pipeline~\cite{torantine2023autogpt,narvekar2020curriculum}, and reliance on expensive server-grade GPUs~\cite{guo2024large}.

Meanwhile, edge computing resources (e.g., RTX 4090, Jetson boards, and Apple M-series) are becoming more powerful and available. These devices provide opportunities to bring ML workloads from cloud cluster to decentralized systems. Although decentralized systems have been studied for robustness and coordination - such as wireless sensor networks~\cite{dimakis2010}, distributed robotics~\cite{oliva2019gossip}, and blockchain~\cite{shi2025hide} - their integration with heterogeneous LLM agents has not been widely addressed.

Thus, a decentralized orchestration runtime which is able to efficiently allocate tasks, operate on heterogeneous devices, and maintain robustness without a centralized orchestrator is urgently needed. In this work, we present \textbf{Symphony}, a decentralized multi-agent system in which lightweight LLMs on edge devices coordinate to achieve intelligence across heterogeneous environments. With three key mechanisms, \textbf{Symphony} is capable of completely decentralized workflow:
\begin{itemize}
\item a \textbf{ledger} which dynamically records device availability and agent capabilities
\item a \textbf{Beacon-based selection protocol} that allocate tasks dynamically and precisely to best-match agents via Beacon~\cite{beacon_distributed_systems}
\item \textbf{weighted result voting} that aggregates results from diverse Chain-of-Thoughts (CoTs)~\cite{wei2022chain}.
\end{itemize}

Together, the mechanisms enable \textbf{Symphony} to orchestrate ML workloads on heterogeneous edge devices in a privacy saving, scalable, and fault-tolerant way.

\section{Methodology}

\textbf{Symphony} is a decentralized multi-agent framework designed for scalable, privacy-preserving collaboration across heterogeneous edge devices. 

\subsection{System Components}

\paragraph{Decentralized Ledger}
A decentralized ledger stores records each agent's capability and availability. 

\paragraph{Worker Nodes}
Worker nodes are edge devices equipped with a quantized LLM(e.g., Mistral-7B~\cite{mistral7binstructv03}), a set of \textbf{stage-specific prompts}, and a lightweight \textbf{Communicator}.

\paragraph{Gateways}
Gateways provide standardized APIs for agent registration, communication, and participation in the task execution process. 

Due to space constraints, the details of system components are provided in Appendix~\ref{app:system_component}.

\subsection{Execution Pipeline}

\begin{figure*}[!ht]
    \centering
    \includegraphics[width=\linewidth]{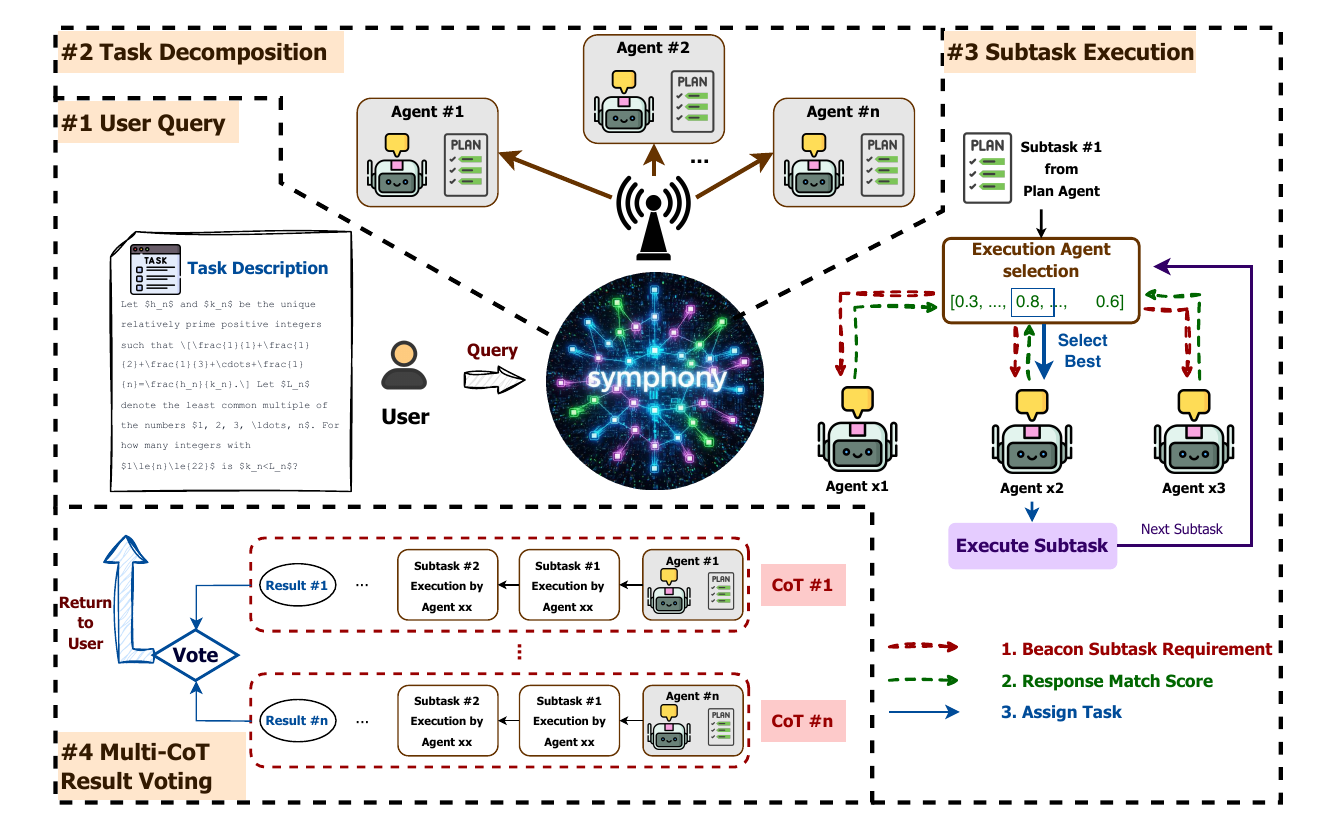}
    \caption{Overview of \textbf{Symphony}. 1. A query from user is decomposed into multiple sub-tasks by planning agents. 2. Sub-task execution leverages Beacon-based agent selection to choose appropriate agents. 3. Final response is generated through result voting across multiple reasoning paths.}
    \label{fig:overview}
\end{figure*}

\paragraph{User Query}
The process begins when the user submits a task description to Symphony. This query is broadcast to multiple planning agents, each responsible for producing a unique CoT.

\paragraph{Planning Phase}
The original description $T_0$ is broadcast to a set of planning agents, each operating independently. Upon receiving $T_0$, a planning agent $E_p$ first extracts the background information $\mathcal{B}$, and then independently generates a decomposition plan composed of a logically ordered sequence of smaller, executable sub-tasks, forming multiple distinct chains-of-thought (CoTs)~\cite{wei2022chain}.

\paragraph{Sub-task Execution}
Suppose there are $M$ CoT $\tau_i$ consisting of of $K_i$ sub-tasks 
$\tau_i = \{t_{i,1}, \dots, t_{i,K_i}\}$. For each sub-task $t_{i,k}$, the plan agent broadcast a \textit{Beacon} $B_{i,k}$~\cite{beacon_distributed_systems} that describes the sub-task requirements to all available agents $\mathcal{E} = \{E_1, \dots, E_N\}$. Upon receiving this beacon $B_{i,k}$, each agent $E_j$ evaluates its own capability vector and computes a \emph{capability match score}
\begin{equation}
    s_{j}(t_{i,k}) = \phi \big( \mathbf{c}_j, \mathbf{r}(t_{i,k}) \big) \in [0,1],
\end{equation}
where $\mathbf{c}_j$ is the capability vector of $E_j$,  
$\mathbf{r}(t_{i,k})$ is the requirement of $t_{i,k}$,  
and $\phi(\cdot,\cdot)$ is a similarity function (e.g., cosine similarity). These scores are then returned to the current agent, which compares them and selects the executor as the agent with the highest score. The selected agent $E_{j^\ast}$ receives the sub-task $t_{i,k}$ along with relevant context from previously completed sub-tasks, executes it locally, and sends the output to the next executor.

\paragraph{Result Voting}
The planning agents independently produce distinct CoTs, each representing a complete reasoning trajectory. After all sub-tasks in $\tau_i$ are executed, the final executer $E_{f}$ outputs a final answer $a_i$ along with an aggregated \emph{confidence score} which is computed as the average of capability match scores along all $K_i$ sub-tasks. Once all CoTs complete execution, their final results are collected into a candidate set $\mathcal{A} = \{a_1, \dots, a_M\}$.  
The final answer $\hat{a}$ is determined by a weighted majority vote:
\begin{equation}
    \hat{a} = \arg\max_{a \in \mathcal{A}} \sum_{i=1}^M \mathbb{I}(a_i = a) \cdot w_i,
\end{equation}
where $\mathbb{I}(\cdot)$ is the indicator function. This voting scheme exploits diversity in reasoning paths to mitigate the impact of individual errors or biases in any single CoT.

\section{Experimental Evaluation}

We evaluate \textbf{Symphony} with the five goals: \textbf{Effectiveness}, \textbf{Scalability across models}, \textbf{Robustness}, and \textbf{Orchestration Overhead}.

\subsection{Experimental Setup}

To reduce demonstrate the advantages of our Symphony framework, we registered three agents in the main experiments. Each task required three distinct chains-of-thought (CoT) via different task decompositions, and final answers were obtained through majority voting. The generation window was set to 512 tokens, with the sampling temperature of 0.5 and a nucleus sampling threshold of $p=0.9$. The decoding process was implemented through the SamplingParams API of the vLLM backend. The runtime environment was composed of three physical servers: Server A equipped with 4 NVIDIA RTX 4090 GPUs (24 GB VRAM), while Server B and C with a single NVIDIA RTX 4090 GPU. All nodes were connected via both public internet and private intranet channels.

Workloads include: (1) \textbf{Big-Bench-Hard}~\cite{srivastava2022beyond}: For each of the 23 types of tasks, 6 questions are randomly chosen, and (2) \textbf{AMC}~\cite{suzgun2022challenging}: 83 competition-style math questions. For Symphony, each task is decomposed into three CoTs, and aggregated via voting.

\subsection{Effectiveness}

\begin{table}[!ht]
\centering
\caption{Accuracy (\%) of Symphony and simplified variants on BBH and AMC benchmarks.}
\renewcommand{\arraystretch}{1.3} 
\resizebox{\columnwidth}{!}{%
\begin{tabular}{llcccc}
\toprule
\hline
\textbf{Benchmark} & \textbf{Model} & \textbf{Direct Solving} & \textbf{AutoGen} & \textbf{CrewAI} & \textbf{Symphony} \\
\midrule
\multirow{3}{*}{\textbf{BBH}} 
 & Deepseek-7B-instruct        & 57.24 & 72.46 & 66.67 & 79.71 \\
\cline{2-6}
 & Mistral-7B-instruct-v0.3    & 36.23 & 48.56 & 50.72 & 78.26 \\
\cline{2-6}
 & Qwen2.5-7B-instruct         & 73.19 & 79.71 & 77.54 & 86.23 \\
\midrule
\multirow{3}{*}{\textbf{AMC}} 
 & Deepseek-7B-instruct        & 10.84 & 8.43 & 7.22 & 13.25 \\
\cline{2-6}
 & Mistral-7B-instruct-v0.3    &  6.02 & 1.79 & 2.40 &  3.61 \\
\cline{2-6}
 & Qwen2.5-7B-instruct         & 16.87 & 21.69 & 18.07 & 25.30 \\
 \hline
\bottomrule
\end{tabular}%
}
\label{tab:model_comparison}
\end{table}

Symphony outperformed LLM-only framework and centralized orchestrations (AutoGen and CrewAI). On BBH, Symphony achieves absolute accuracy gains ranging from 6.5\% to 41.6\% compared with Direct Solving, and 6.5\% to 29.1\% over AutoGen, showing that dynamic decentralized orchestration yields significant improvements even over existing multi-agent frameworks. 

On AMC, which is a more challenging benchmark with lower absolute scores, Symphony still surpasses all baselines, achieving up to 4.46\% higher accuracy than AutoGen and up to 7.41\% higher than Direct Solving. These results show that this decentralized orchestration is not only feasible but also more effective than centralized ones.

\subsection{Scalability across models}

We evaluate \textbf{Symphony} across three LLMs: Deepseek-7B-instruct, Mistral-7B-instruct-v0.3, and Qwen2.5-7B-instruct. As shown in Table~\ref{tab:model_comparison}, Symphony benefits all LLMs across the two benchmarks. Meanwhile, while direct solving demonstrate a wide accuracy gap among LLMs (36\%–73\% on BBH), with Symphony, the accuracy gap narrows to 78\% - 87\%. This indicates that Symphony particularly enhances weaker models, demonstrating its potential capability on heterogeneous devices.

\begin{table*}[!ht]
\centering
\begin{minipage}{0.48\linewidth}
\centering
\caption{Effect of CoT Voting: Comparison between Single CoT and 3-CoT Voting}
\label{tab:cot_voting_ablation}
\renewcommand{\arraystretch}{1.2}
\resizebox{\linewidth}{!}{
\begin{tabular}{llccc}
\toprule
\textbf{Benchmark} & \textbf{Model} & \textbf{1 CoT} & \textbf{3 CoT Voting} & \textbf{Improvement} \\
\midrule
BBH & Deepseek-7B & 75.36 & 79.71 & +4.25 \\
    & Mistral-7B & 71.74 & 78.26 & +6.52 \\
    & Qwen2.5-7B & 81.16 & 86.23 & +5.07 \\
AMC & Deepseek-7B & 11.45 & 13.25 & +1.80 \\
    & Mistral-7B &  2.89 &  3.61 & +0.72 \\
    & Qwen2.5-7B & 22.67 & 25.30 & +2.63 \\
\bottomrule
\end{tabular}}
\end{minipage}\hfill
\begin{minipage}{0.48\linewidth}
\centering
\caption{Effect of Beacon Score-based Selection: Random vs. Score Selection}
\label{tab:beacon_selection_ablation}
\renewcommand{\arraystretch}{1.2}
\resizebox{\linewidth}{!}{
\begin{tabular}{llccc}
\toprule
\textbf{Benchmark} & \textbf{Model} & \textbf{Random} & \textbf{Score} & \textbf{Improvement} \\
\midrule
BBH & Deepseek-7B & 76.09 & 79.71 & +3.62 \\
    & Mistral-7B & 73.91 & 78.26 & +4.35 \\
    & Qwen2.5-7B & 82.61 & 86.23 & +3.62 \\
AMC & Deepseek-7B & 11.85 & 13.25 & +1.40 \\
    & Mistral-7B &  3.01 &  3.61 & +0.60 \\
    & Qwen2.5-7B & 23.12 & 25.30 & +2.18 \\
\bottomrule
\end{tabular}}
\end{minipage}
\end{table*}

\subsection{Robustness}

Ablation experiments were conducted to study the impact of Beacon Selection and CoT Voting. As shown in Table~\ref{tab:cot_voting_ablation},the multi-CoT voting mechanism improves performance across all models, with BBH gains of +5.3\% to +6.2\% and AMC gains of +0.72\% to +2.63\%. Table~\ref{tab:beacon_selection_ablation} illustrates that the Beacon score-based selection consistently outperforms random allocation, with BBH gains of +4.1\% to +4.3\% and AMC gains of +0.60\% to +2.18\%. Both of the mechanisms serve as robustness enhancers for Symphony. Multi-CoT voting improves tolerance of failure, while Beacon Selection guarantees that subtasks are allocated to best-matches.

\subsection{Orchestration Overhead}

We measured the end-to-end overhead introduced by Symphony's mechanisms, including ledge registration, beacon broadcast, and result voting. Across all evaluated tasks, these process together contributed less than 5\% of the inference latency. This demonstrate that Symphony's orchestration cost is negligible compared with model inference time. 

\section{Conclusion}
We presented \textbf{Symphony}, a decentralized multi-agent framework that enables scalable collaboration across heterogeneous edge devices through self-play, sparse parameter sharing, and role-specific cooperation. Experiments show that Symphony achieves competitive performance with significantly lower communication and infrastructure costs, while enhancing accessibility, preserving privacy, and supporting the emergence of decentralized agent economies.

\bibliographystyle{unsrtnat}
\bibliography{reference}

\appendix

\section{System Component}
\label{app:system_component}

\paragraph{Decentralized Ledger}
A decentralized ledger stores each agent’s resource ownership, contribution records, and domain expertise, indexed by a DID-compliant cryptographic address. 

\paragraph{Worker Nodes}
Worker nodes are edge devices with full local autonomy, such as consumer-grade GPUs or Apple M-series machines. Each node integrates three key components. The first is the \textbf{Local Engine}, a quantized LLM (e.g., Mistral-7B~\cite{mistral7binstructv03}) optimized for on-device inference to reduce latency and resource consumption. The second is a set of \textbf{stage-specific prompts} that are automatically selected according to the current phase of the execution pipeline, ensuring that planning, sub-task execution, and result aggregation each receive context-appropriate instructions. The third is the \textbf{Communicator}, a lightweight and secure messaging module that enables efficient inter-agent communication over both intranet and public networks while maintaining data privacy.

\paragraph{Gateways}
Gateways provide standardized APIs for agent registration, communication, and participation in the task execution process. Registration is completed via the user's configuration, requiring specification of model path, GPU allocation, host and post. Messages exchanged among agents are divided into four categories: \textbf{Beacon}, \textbf{Beacon Response}, \textbf{Task}, and \textbf{Task Result}. Upon receiving a message, the agent will automatically invoke the corresponding message-handling function according to its type.

\section{Case Study Details}
\label{app:case_study}

To provide detailed insights into Symphony, we demonstrate the operation pipeline of the following case, which is a casual judgement question from Big-Bench-Hard.

\subsubsection{Task Setup} 
The original task description is :

\begin{quote}
How would a typical person answer each of the following questions about causation?  
Drew, Kylie, Oliver, and Jen are regular customers at a small, local coffee shop. Given the selling price of the coffee and the cost of daily operation, the coffee shop will turn a profit if anyone orders coffee on a given day. Only one person ordering coffee is needed for the coffee shop to turn a profit that day. Kylie and Oliver usually order coffee on Tuesdays. However, Drew doesn't usually order coffee on Tuesdays. This Tuesday, unexpectedly, Drew ordered coffee. The same day, Kylie ordered coffee, and Oliver also ordered coffee. Since at least one person ordered coffee on Tuesday, the coffee shop made a profit that day. Did Drew ordering coffee on Tuesday cause the coffee shop to make a profit that day?
\end{quote}

\noindent \textbf{Options:}
\begin{itemize}
    \item Yes
    \item No
\end{itemize}

This is a multi-choice question of daily life issue, requiring context understanding and logistic capability. 

\subsubsection{Background Extraction and Task Decomposition} 

\begin{figure}[!ht]
    \centering
    \includegraphics[width=0.5\textwidth]{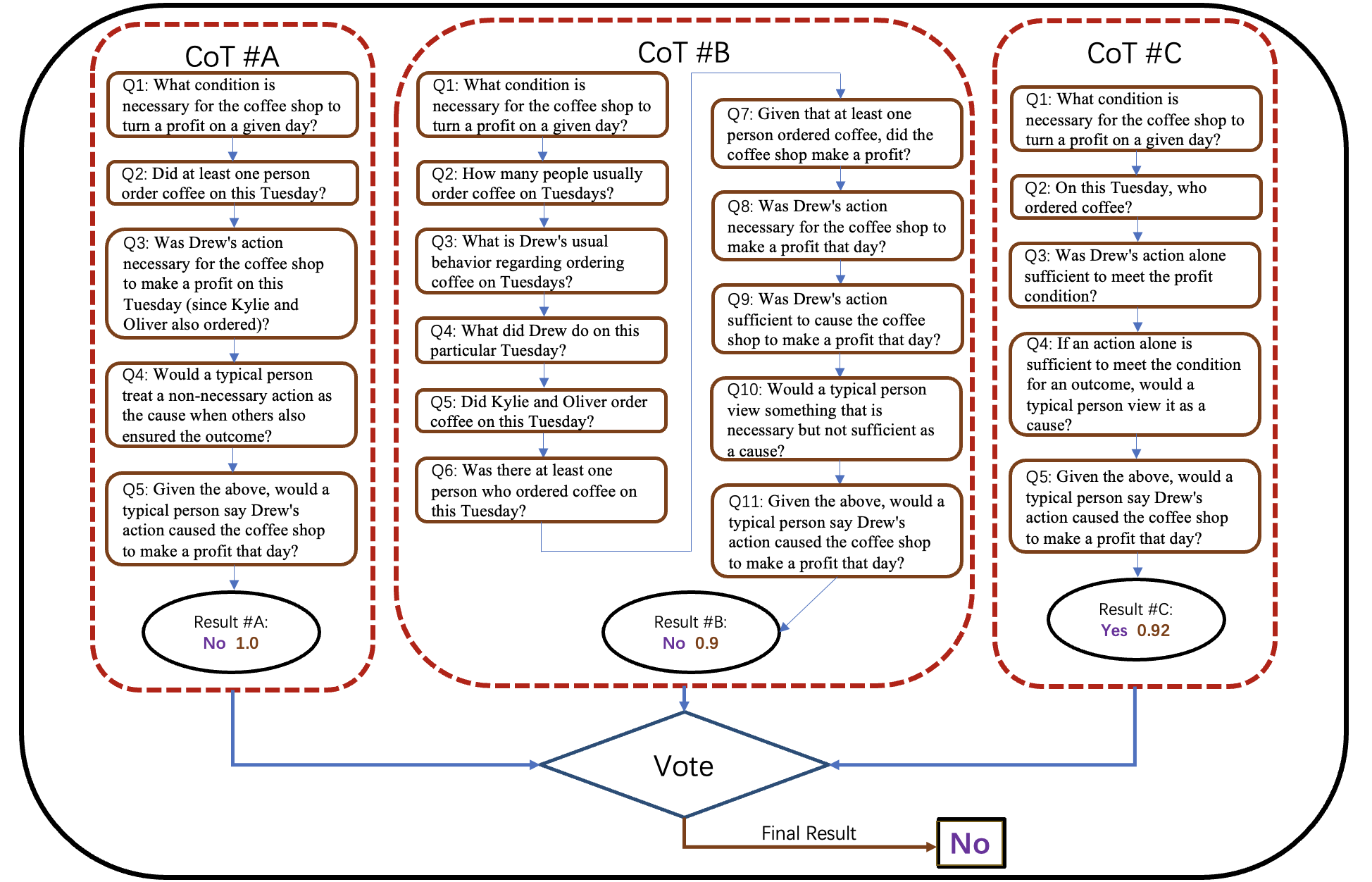}
    \caption{Illustration of the \textbf{Symphony} pipeline on a BBH case. Three independent planning agents generate different CoTs to enhance diversity of solutions.}
    
    \label{fig:case_study}
\end{figure}

In the experiment, 3 CoT were required, so 3 agents with planning capability were chosen from all 8 agents based on capability match and response speed. As shown in Figure~\ref{fig:case_study}, upon receiving the original task description, the 3 plan agents extract the background information, and then decompose the task into Chain-of-Thought independently, ensuring diversity of solutions.

\subsubsection{Sequential Cooperation in Execution}

Unlike isolated single-agent learning, \textbf{Symphony} generates collaborative intelligence through sequential cooperation across agents. When assigned a sub-task, the executor also receives the background information and results of previous sub-tasks, ensuring continuity of reasoning and alignment of thoughts across agents.

For instance, in CoT A, the second sub-task (\textit{Did at least one person order coffee on this Tuesday?}) is executed by a logic-specialized agent. Its output “Yes”, along with the result of sub-task 1, is then passed to the following executors in structured context. Based on the previous results, the third sub-task (\textit{Was Drew's action necessary for the coffee shop to make a profit on this Tuesday (since Kylie and Oliver also ordered)?}) was answered. This chaining process guarantees that every answer is not isolated, but co-constructed by multiple agents.

\subsubsection{Result Aggregation}
The three CoTs produce final answers with a confidential score:

\begin{align*}
\text{CoT A} &\;\;\rightarrow\;\; \text{No} \; (1.0) \\
\text{CoT B} &\;\;\rightarrow\;\; \text{No} \; (0.9) \\
\text{CoT C} &\;\;\rightarrow\;\; \text{Yes} \; (0.92)
\end{align*}

Through weighted majority voting, Symphony aggregates the final result: No. The aggregation process improves the stability of the final answer, avoiding the negative influence of single point failure.

\section{Prompt Design}
\textbf{Symphony} employs two core prompts to implement problem decomposition and solving in different agents. The first prompt is responsible for breaking down complex problems into sequences of computable subtasks, used in planning phase, while the second prompt handles the execution of specific subtask solving, used in sub-task execution phase.

\section{Deployment and Societal Implications}
\label{app:social_implications}

\textbf{Symphony} advances existing multi-agent LLM frameworks such as AutoGen~\cite{autogen2023} and MetaGPT~\cite{metagpt2023} by introducing a decentralized architecture that enables scalable and resilient deployment. Beyond technical improvements, this design carries broader implications for AI accessibility, privacy preservation, and the emergence of agent-based economies.

A key advantage of \textbf{Symphony} is its ability to lower hardware requirements to consumer-grade GPUs, thereby reducing reliance on centralized cloud infrastructure. This capability empowers individuals, small teams, and communities with limited computational resources to participate in collaborative intelligence creation, significantly lowering entry barriers. For example, a local medical research group in a developing region could deploy multiple lightweight agents on existing desktop GPUs to analyze anonymized radiology datasets, collaboratively generating diagnostic insights without uploading sensitive patient data to external servers. Such deployment not only circumvents the high costs of cloud computing but also ensures compliance with local data governance and privacy regulations.

The framework’s decentralized paradigm also inherently strengthens privacy guarantees. Task execution remains confined to local devices, and only concise sub-task outcomes are broadcast within the network. This ensures that sensitive information never leaves local storage. In a cross-hospital medical AI collaboration, for instance, each participating hospital could run Symphony agents locally to process patient imaging data and extract intermediate diagnostic features. These features, stripped of personally identifiable information, would then be shared within the network for joint reasoning, enabling collaborative model improvements and consensus diagnosis without exposing raw patient records. This approach preserves data sovereignty, satisfies regulatory requirements such as HIPAA~\cite{hipaa1996} or GDPR~\cite{gdpr2016}, and maintains high diagnostic performance.

Finally, \textbf{Symphony} enables decentralized agent economies through agent-level autonomy, supporting independent decision-making, local evaluation, and incentive-driven collaboration. This aligns with economic models in which agents behave as rational participants—bidding for tasks, providing services, and adapting strategies. For example, in a global open-source software development network, independent Symphony agents operated by different contributors could autonomously bid for programming tasks, such as implementing a feature or fixing a bug, based on their historical performance and domain expertise. Successful completion would yield digital tokens or credits that could be exchanged for compute resources, premium datasets, or other services within the ecosystem. Such a structure fosters a self-sustaining, market-like environment where agents continuously adapt to changing demands and resource availability.

\begin{promptbox}{Subtask Execution Prompt}{prompt:execution}
You are given background information, including previous questions and answers, as well as relevant context. Based on this context, solve the current sub-task and provide the final answer formatted as $\boxed{<Answer>}$. Do not provide additional explanations or code.

Here are some examples:

Example 1:
Input: Background information include: "Consider two positive even integers less than $15$ (not necessarily distinct)". Based on the background information, solve the sub-task: "What are the possible values for the two positive even integers less than 15?". Provide the final answer formatted as $\boxed{<Answer>}$. Do not provide additional explanations or code.

Output: $\boxed{2, 4, 6, 8, 10, 12, 14}$

Example 2:
Input: Background information include: "If $23=x^4+\frac{1}{x^4}$. Q1: How can we express $x^4 + \frac{1}{x^4}$ in terms of $x^2 + \frac{1}{x^2}$? Answer: $\boxed{(x^2 + \frac{1}{x^2})^2 - 2}$". Based on the background information, solve the sub-task: "Q2: Given that $23 = x^4 + \frac{1}{x^4}$, what is the value of $x^2 + \frac{1}{x^2}$?". Provide the final answer formatted as $\boxed{<Answer>}$. Do not provide additional explanations or code.

Output: $\boxed{5}$

DO NOT provide additional explanations or code.

Here is the current sub-task:

Input: Background information include: "{context}". 
Based on the background information, solve the sub-task: "{instruction}". 
Provide the final answer formatted as $\boxed{{<Answer>}}$. 
Do not provide additional explanations or code. Output:
\end{promptbox}

\begin{promptbox}{Problem Decomposition Prompt}{prompt:decomposition}
You are a problem decomposer, not a solver. Your task is to break down a complex math or logic problem into a sequence of strictly computable sub-questions. Each sub-question must represent a well-defined, executable step toward solving the original problem.

Each subtask must be phrased as a question. Do not solve the problem or output the final answer.

You MUST strictly output the result in the following **valid JSON** format:

Output:
{
"original_question": "<repeat the original question>",
"subtasks": [
    "Q1: ...",
    "Q2: ...",
    ...
]
}

Important Rules:

- Do NOT include any final answer, intermediate answer, or numerical result.
- Do NOT perform or explain any computation.
- Do NOT include any text outside the JSON object.
- Each subtask must be directly computable (e.g., calculate a value, rewrite an expression, identify a condition).
- Use clear and concise language appropriate for step-by-step problem solving.

Here are some examples:

Example 1:
Input: One root of the equation $5x^2+kx=4$ is 2. What is the other?

Output:
{
  "original_question": "One root of the equation $5x^2+kx=4$ is 2. What is the other?",
  "subtasks": [
    "Q1: What is the equation rewritten in standard quadratic form?",
    "Q2: What is the product of the roots of this quadratic equation?",
    "Q3: Given one root is 2, what is the other root?"
  ]
}

Do NOT include any explanation, prefix, or suffix before or after the JSON. Output ONLY the JSON object.

Now decompose the following problem:

Input: {user_input}

Output:
\end{promptbox}

\end{document}